\documentclass[journal]{IEEEtran}
\usepackage{amsmath,amsfonts}
\usepackage{algorithmic}
\usepackage{algorithm}
\usepackage{array}
\usepackage[caption=false,font=footnotesize,labelfont=rm,textfont=rm]{subfig}
\usepackage{xcolor}
\usepackage[normalem]{ulem}

\usepackage{textcomp}
\usepackage{stfloats}
\usepackage{url}
\usepackage{verbatim}
\usepackage{graphicx}
\usepackage{makecell}
\usepackage{cite}
\usepackage{booktabs}  
\usepackage{multirow}  
\usepackage[colorlinks,linkcolor=blue,anchorcolor=blue,citecolor=blue,]{hyperref}
\graphicspath{{./img/}}

\begin{document}

\title{Scenario-Adaptive MU-MIMO OFDM Semantic Communication With Asymmetric Neural Network}

\author{Chongyang Li, Tianqian Zhang, and Shouyin Liu
\thanks{Chongyang Li, Tianqian Zhang, and Shouyin liu are with the Department of
	Electronics and Information Engineering, the College of Physical Science and Technology, Central China Normal University, Wuhan 430079, China (email: lichongyang2020@mails.ccnu.edu.cn; tqzhang@mails.ccnu.edu.cn; syliu@ccnu.edu.cn)}
	
\thanks{This work was supported in part by the Key Research and Development Program of Hubei Province under Grant 2023BAB061, and in part by the Natural Science Foundation of Hubei Province under Grant 2025AFB327, and in part by China Postdoctoral Science Foundation under Grant 2023M741315, and in part by Postdoctor Project of Hubei Province under Grant 2024HBBHCXA049.}

\thanks{Our project and open source code are available at: \url{https://github.com/Linkcy97/MUMIMOSC}}
}



\maketitle

\begin{abstract}
Semantic Communication (SemCom) has emerged as a promising paradigm for 6G networks, aiming to extract and transmit task-relevant information rather than minimizing bit errors. However, applying SemCom to realistic downlink Multi-User Multi-Input Multi-Output (MU-MIMO) Orthogonal Frequency Division Multiplexing (OFDM) systems remains challenging due to severe Multi-User Interference (MUI) and frequency-selective fading. Existing Deep Joint Source-Channel Coding (DJSCC) schemes, primarily designed for point-to-point links, suffer from performance saturation in multi-user scenarios. To address these issues, we propose a scenario-adaptive MU-MIMO SemCom framework featuring an asymmetric architecture tailored for downlink transmission. At the transmitter, we introduce a scenario-aware semantic encoder that dynamically adjusts feature extraction based on Channel State Information (CSI) and Signal-to-Noise Ratio (SNR), followed by a neural precoding network designed to mitigate MUI in the semantic domain. At the receiver, a lightweight decoder equipped with a novel pilot-guided attention mechanism is employed to implicitly perform channel equalization and feature calibration using reference pilot symbols. Extensive simulation results over 3GPP channel models demonstrate that the proposed framework significantly outperforms DJSCC and traditional Separate Source-Channel Coding (SSCC) schemes in terms of Peak Signal-to-Noise Ratio (PSNR) and classification accuracy, particularly in low-SNR regimes, while maintaining low latency and computational cost on edge devices.
\end{abstract}

\begin{IEEEkeywords}
Semantic Communication, Multi-User MIMO, OFDM, deep learning, neural network
\end{IEEEkeywords}

\section{Introduction}
\IEEEPARstart{T}{h}e past few decades have witnessed an unprecedented evolution in wireless communication systems, transitioning from the analog voice transmission of first generation (1G) to the high-speed, low-latency connectivity of fifth generation (5G). The 5G base stations utilize massive Multiple-Input Multiple-Output (MIMO) \cite{lu2014overview}, millimeter wave (mmWave), and ultra-dense networking (UDN) to support up to 64 transceiver chains with increased antenna elements\cite{wang2023road}. With the rise of artificial intelligence (AI), the sixth generation (6G) is expected to usher in a new paradigm of AI-assisted wireless communication. Specifically, conventional communication paradigms have primarily focused on the reliable transmission of bit sequences, evaluating system performance through metrics such as bit error rate (BER) and data throughput. In contrast, the emerging paradigm of Semantic Communication (SemCom) leverages deep learning to extract and transmit the underlying meaning or task-relevant features of the data, thereby shifting the design objective from bit-level exactness to semantic fidelity and task execution efficiency.

\begin{figure*}[ht]
	\centering
	\includegraphics[width=6in]{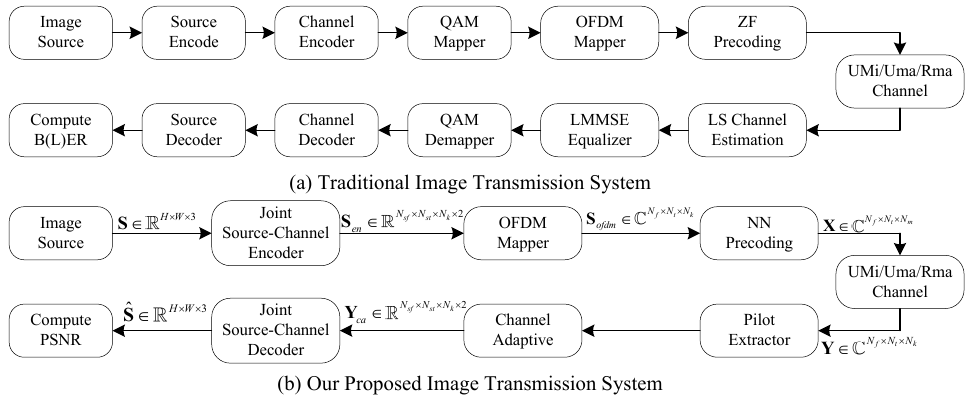}
	\caption{Comparison of Our Proposed Image Transmission System with Traditional Image Transmission Systems}
	\label{sys}
\end{figure*}

The design of traditional systems has been governed by Shannon’s separation theorem \cite{shannon1948mathematical}, which advocates Separate Source–Channel Coding (SSCC). As illustrated in Fig. \ref{sys}(a), this paradigm treats source and channel coding as independent modules optimized specifically for bit-level reliability. Although conventional modular architectures approach the Shannon limit, they struggle to handle diverse and large-scale traffic in emerging 6G networks.

To break through these physical constraints, AI-assisted SemCom has emerged as a revolutionary paradigm for 6G, aiming to realize the semantic and effectiveness levels of communication originally envisioned by Weaver and Shannon \cite{shannon_weaver_1949}. Unlike traditional systems that focus on symbol reproduction, SemCom prioritizes the precise conveyance of meaning and the effectiveness of information in serving downstream tasks. This paradigm shift is predominantly realized through Deep Joint Source-Channel Coding (DJSCC) \cite{bourtsoulatze2019deep}, which leverages deep neural networks to extract compact semantic features and map them directly to channel symbols in an end-to-end manner. By integrating source and channel coding into a unified differentiable framework, DJSCC significantly enhances transmission efficiency and robustness under resource constrained wireless environments.

SemCom has demonstrated remarkable versatility across various modalities, including text\cite{xie2021deep}, speech\cite{weng2021semantic}, and images\cite{huang2022toward}. The unifying principle in these applications is the extraction and conveyance of semantic essence, relaxing the constraints on bit-level precision. For instance, text transmission safeguards linguistic coherence; speech transmission focus on the integrity of the message or speaker characteristics; and image transmission prioritizes the retention of visual content and structure. Given that traditional BER metrics fail to capture such semantic quality, evaluation protocols have evolved to include modality specific criteria. These include Bilingual Evaluation Understudy (BLEU) \cite{papineni2002bleu} for text generation, Perceptual Evaluation of Speech Quality (PESQ) \cite{rix2001perceptual} for speech quality, and Peak Signal-to-Noise Ratio (PSNR) alongside Multi-Scale Structural Similarity Index Measure (MS-SSIM) \cite{wang2003multiscale} for image reconstruction fidelity.

However, despite these promising advancements, Most existing research on image SemCom relies on simplified physical layer models, such as Additive White Gaussian Noise (AWGN), which fail to capture the complexity of practical deployment scenarios \cite{yang2023witt,xu2023deep,peng2025robust,li2025mvsc}. Although some recent studies have integrated MIMO and Orthogonal Frequency Division Multiplexing (OFDM) technologies to address multipath fading\cite{yang2022ofdm,wu2024deep,liang2025semantic,liu2025semantic}, the majority of these works are confined to single user point-to-point links. This simplification fails to capture the practical characteristics of modern cellular networks, where the Base Station (BS) simultaneously serves multiple users via downlink Multi-User MIMO (MU-MIMO). Consequently, extending semantic communication from P2P to MU-MIMO in realistic downlink scenarios remains largely underexplored.  

Moreover, there are two critical problems often ignored in practical deployment. First is the hardware asymmetry. In downlink scenario, the BS possesses abundant computational resources and power, whereas User Equipment (UE), such as Internet of Things (IoT) sensors or mobile devices, is strictly constrained by battery life and processing power. The symmetric autoencoder architectures commonly adopted in prior work\cite{yang2024swinjscc,bourtsoulatze2019deep,xu2023deep,peng2025robust,yang2022ofdm,wu2024deep,liang2025semantic,liu2025semantic}, in which the decoder has a complexity comparable to that of the encoder, are unsuitable for lightweight UE. Second is the environmental adaptability. Real world channels vary drastically between Urban Micro (UMi), Urban Macro (UMa), and Rural Macro (RMa) scenarios, characterized by different path losses and delay spreads. A model trained for a specific channel statistic typically fails to generalize, and training separate models for every scenario is inefficient.

To address these challenges, this paper proposes a novel scenario-adaptive SemCom framework feature an asymmetric encoder-decoder architecture for downlink MU-MIMO OFDM systems.

The main contributions of this paper are summarized as follows:

\begin{enumerate}
	\item  Semantic Downlink MU-MIMO Architecture: We propose a pioneering SemCom framework tailored for downlink MU-MIMO OFDM systems. This architecture enables concurrent semantic transmission to multiple users by jointly optimizing feature extraction and neural precoding, thereby ensuring high semantic fidelity in complex multi-user environments.
	
	\item Neural Precoding: We develop a robust neural precoding module to replace traditional linear precoders within the training loop. This data-driven approach eliminates numerical instabilities associated with matrix inversion of singular channels, effectively preventing gradient explosion and ensuring robust end-to-end convergence.
	
	\item Asymmetric Encoder-Decoder Design: Addressing the processing capability disparity between base stations and user equipment, we design an asymmetric encoder-decoder architecture. The encoder incorporates deep residual attention blocks for robust feature extraction, whereas the decoder utilizes depthwise separable convolutions to minimize computational overhead and parameter count while maintaining reconstruction quality.
	
	\item Spectrum-Aware Scenario Adaptivity: An attention feature modulation mechanism is introduced to adaptively recalibrate feature importance based on Channel State Information (CSI). By analyzing the magnitude spectrum to infer path loss and delay spread characteristics, this module enables the system to generalize effectively across diverse propagation scenarios such as UMi, UMa, and RMa.
	
	\item Pilot-Guided Attention: We implement a pilot-guided attention decoding strategy that concatenates received pilot signals and corresponding known reference symbols into the reconstruction network. Unlike implicit schemes, this approach explicitly leverages CSI to enhance signal restoration and image reconstruction performance.
	
\end{enumerate}

\begin{figure}[ht]
	\centering
	\includegraphics[width=3in]{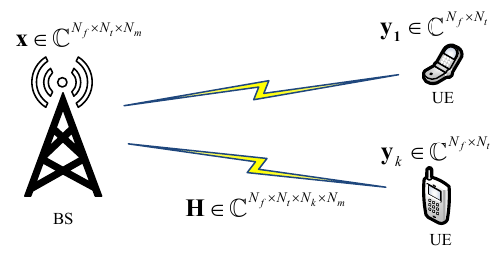}
	\caption{The comparison of three different network blocks}
	\label{fig_2}
\end{figure}

\textit{Notations:} $\mathbb{R}$ and $\mathbb{C}$ denote the sets of real and complex numbers, respectively. Tensors and matrices are denoted by bold upper-case letters, while vectors are represented by bold lower-case letters. Scalars are denoted by italic letters. For a 4D tensor $\mathbf{H} \in \mathbb{C}^{N_a \times N_b \times N_c \times N_d}$, we denote by $\mathbf{H}_{a,b} \in \mathbb{C}^{N_c \times N_d}$ the matrix, and $\mathbf{h}_{a,b,c} \in \mathbb{C}^{N_d}$ the vector formed by slicing $\mathbf{H}$ along the first two and three dimensions, respectively. $(\cdot)^{H}$ denotes the  Hermitian matrix. $\mathbf{I}$ is the identity matrix. $\mathbb{E}[\cdot]$ denotes the statistical expectation operation.

\section{System Model}\label{System}
We consider a downlink Multi-User MIMO (MU-MIMO) system, where a BS equipped with $N_m$ transmit antennas serves $N_k$ UE as shown in Fig. \ref{fig_2}. Each UE is equipped with a single receive antenna. Considering an OFDM resource grid consisting of $N_f$ subcarriers and $N_t$ OFDM symbols, the channel coefficients form a 4-dimensional tensor denoted by $\mathbf{H} \in \mathbb{C}^{N_f \times N_t \times N_k \times N_m}$. Consequently, $\mathbf{H}_{f,t} \in \mathbb{C}^{N_k \times N_m}$ represents the channel matrix at resource element (RE) $(f,t)$. Specifically, the row vector $\mathbf{h}_{k,f,t} \in \mathbb{C}^{1 \times N_m}$ denotes the channel from the BS to user $k$. The received signal vector $\mathbf{y}_{f,t}$ at is given by:
 
\begin{equation}
	\mathbf{y}_{f,t} = \mathbf{H}_{f,t}\mathbf{x}_{f,t} + \mathbf{n}_{f,t}
\end{equation}

where $\mathbf{x}_{f,t} \in \mathbb{C}^{N_m} $ is the transmitted signal and $n_{f,t} \sim \mathcal{CN}(0, \sigma^2)$ is the additive white Gaussian noise. For a specific user $k$, the received scalar $y_{k,f,t}$ is expressed as:

\begin{equation}
	y_{k,f,t} = \mathbf{h}_{k,f,t} \mathbf{v}_{k,f,t} s_{k,f,t} + \sum_{j \neq k} \mathbf{h}_{j,f,t} \mathbf{v}_{j,f,t} s_{j,f,t} + n_{j,f,t}
	\label{eq:system_model}
\end{equation}

where $\mathbf{v}_{k,f,t} \in \mathbb{C}^{N_m \times 1}$ represents the precoding vector that spatially maps the semantic symbol onto the $N_m$ transmit antennas. The term $s_{k,f,t} \in \mathbb{C}$ denotes the normalized semantic feature symbol for user $k$. The first term in (\ref{eq:system_model}) represents the desired semantic signal enhanced by beamforming gain, while second term explicitly characterizes the MUI caused by the superposition of precoded signals intended for other users $j \neq k$.

To suppress such interference, linear precoding schemes are conventionally employed. A representative method is the Regularized Zero-Forcing (RZF) precoding \cite{peel2005vector}, which balances interference mitigation and noise enhancement. The RZF precoding matrix $\mathbf{V}_{f,t} = [\mathbf{v}_{1,f,t}, \ldots, \mathbf{v}_{k,f,t}] \in \mathbb{C}^{N_m \times N_k}$ is typically computed as:

\begin{equation} 
	\mathbf{V}_{f,t} = \mathbf{H}_{f,t}^H \left( \mathbf{H}_{f,t}\mathbf{H}_{f,t}^H + \alpha \mathbf{I} \right)^{-1}, 
	\label{eq:rzf} 
\end{equation}
where $(\cdot)^H$ denotes the conjugate transpose, $\alpha$ is a regularization factor with typical choices such as noise variance. If $\alpha=0$, it degenerates into ZF precoding. 

The overall transmission procedure of the proposed system is illustrated in Fig. \ref{sys}(b). At the transmitter, the source image $\mathbf{S} \in \mathbb{R}^{H \times W \times 3}$ is first processed by a joint source–channel encoder, which directly maps visual content into compact semantic features, expressed as 
\begin{equation}
	\mathbf{S}_{en}=f_e(\mathbf{S};\mathbf{\phi}).
\end{equation}
The encoder outputs a real number $\mathbf{S}_{en} \in \mathbb{R}^{N_{sf} \times N_{st} \times N_k \times 2}$, where $N_{sf}$ and $N_{st}$ denote the number of subcarriers and symbols prior to pilot insertion. The last dimension of size two  represents the real and imaginary components, as neural networks typically operate on real number tensors. These features are then mapped onto the physical resource grid and augmented with pilot sequences to form the complex number symbols $\mathbf{S}_{odfm} \in \mathbb{C}^{N_f \times N_t \times N_k}$. Subsequently, a neural network precoder exploits the spatial degrees of freedom of the MU-MIMO channel to generate the transmitted signal $\mathbf{X} \in \mathbb{C}^{N_f \times N_t \times N_m}$. This precoder optimizes the mapping of semantic symbols onto the $N_m$ transmit antennas to effectively  mitigate Multi-User Interference (MUI). The precoded signals are transmitted over 3GPP wireless channels.

At the receiver side, the transmitted signal undergoes channel fading and additive noise, yielding the received signal $\mathbf{Y} \in \mathbb{C}^{N_f \times N_t \times N_k}$. Pilot symbols embedded in the OFDM frame are explicitly extracted to facilitate channel-aware feature recovery. Utilizing both the received signal and the extracted pilot information, a channel-adaptive module performs feature calibration and distortion compensation. This module then removes the pilot to output the refined semantic features $\mathbf{Y_{ca}} \in \mathbb{R}^{N_{sf} \times N_{st} \times N_k \times 2}$. These features are then fed into a lightweight joint source–channel decoder to reconstruct the image $\mathbf{\hat{S}} \in \mathbb{R}^{H \times W \times 3}$, defined as
\begin{equation}
	\mathbf{\hat{S}}=f_d(\mathbf{Y_{ca}};\mathbf{\theta}).
\end{equation}
The encoder and decoder are jointly trained in an end-to-end manner by minimizing a reconstruction loss function. Specifically, the learning objective can be formulated as
\begin{equation}
	\min_{\phi, \theta} \mathbb{E}_{\mathbf{S} \sim p_\mathbf{S}, \; \mathbf{\hat{Y}} \sim p_{\mathbf{\hat{Y}}|\mathbf{S}}} \left[ \mathcal{L}(\mathbf{S}, f_{\mathrm{dec}}(\mathbf{\hat{Y}}; \theta)) \right],
\end{equation}
where $\mathcal{L}(\cdot, \cdot)$ represents a distortion measure that quantifies the discrepancy between the original image and its reconstruction. For image transmission tasks, the reconstruction performance is typically evaluated using metrics such as PSNR and MS-SSIM, which characterize the reconstruction quality from pixel-level fidelity and structural perceptual consistency perspectives. Owing to the explicit pilot-guided attention and the asymmetric encoder–decoder design, the proposed end-to-end framework achieves robust image transmission under diverse channel conditions while maintaining low receiver complexity. 

To evaluate the proposed system under realistic wireless propagation conditions, we adopt standardized channel models specified in 3GPP TR 38.901 \cite{3GPP_TR38901_V18}. Specifically, we focus on three representative outdoor deployment scenarios: UMa, UMi, and RMa. These scenarios are chosen as they represent the most critical and diverse environments for outdoor downlink MU-MIMO communications. By encompassing both Line-of-Sight (LOS) and Non-LOS (NLOS) propagation conditions , this selection provides a comprehensive evaluation covering a wide range of path loss, delay spread, and severe multipath fading characteristics.

By incorporating these standardized channel models into the end-to-end learning framework, the proposed system is trained and evaluated across diverse propagation environments, enabling it to adapt effectively to different channel characteristics and deployment scenarios.

\begin{figure*}[ht]
	\centering
	\includegraphics[width=6in]{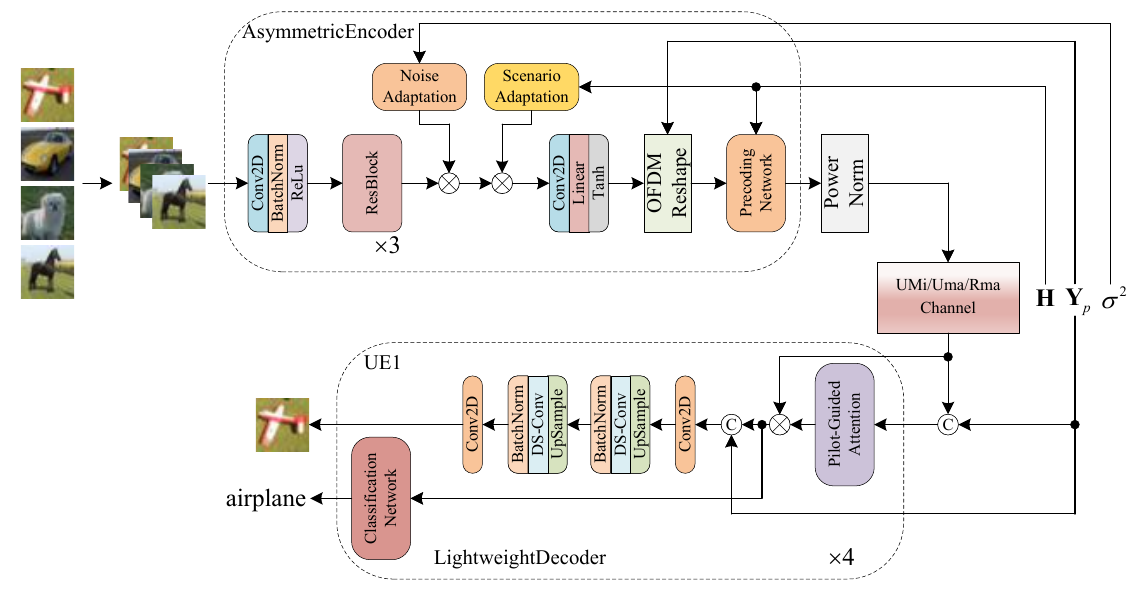}
	\caption{The overall architecture of the proposed scheme for wireless image transmission.}
	\label{fig_3}
\end{figure*}

\section{The proposed framework}\label{framework}
In this section, we introduce the proposed MU-MIMO OFDM SemCom framework as shown in Fig. \ref{fig_3}. At the BS, the input images from all users are first collected and processed by an asymmetric semantic encoder. The encoder extracts high-level semantic features from each image using deep convolutional layers, transforming raw visual data into compact latent representations. To enhance robustness under diverse propagation conditions, two dedicated adaptation mechanisms are incorporated.

Specifically, a scenario adaptation module exploits large-scale channel characteristics to adjust semantic feature representations according to channel frequency responses, while a noise adaptation module leverages the noise variance to modulate feature amplitudes, enabling resilience against varying noise levels. In this work we assumed that the channel responses and noise variance are known at the transmitter.

After adaptation, the encoded features of all users are mapped onto the OFDM resource grid, where time-frequency RE are explicitly organized to support multi-user transmission, and known pilot symbols are embedded at predefined locations. A learnable precoding network is then applied to the mapped symbols, jointly considering  multi-user interference and spatial channel characteristics. This network generates precoded signals across multiple transmit antennas, enabling spatial multiplexing in the downlink. Finally, a power normalization operation is performed to ensure compliance with the average transmit power constraint before transmission over realistic UMi, UMa, or RMa channels.

At the receiver side, each UE independently processes its received signal. The received symbols are first concatenated with the corresponding pilot and fed into a pilot-guided module, which learns a channel-adaptive modulation map from the known pilot reference. The output of the pilot-guided module is then applied to the received symbols through element-wise multiplication, enabling explicit channel-aware feature refinement. Subsequently, the modulated features are concatenated with the known pilot information in a residual manner. 

The refined features are subsequently fed into two parallel branches at each UE. One branch is an image reconstruction network, which aims to recover the original image by minimizing the mean squared error (MSE) loss. The other branch is an image classification network, which directly infers semantic labels from the received features and is trained using the cross-entropy (CE) loss. This design enables simultaneous support of pixel-level reconstruction and task-oriented semantic inference within a unified framework. All UEs perform decoding in parallel using identical lightweight decoder architectures.

The detailed network architecture of the proposed SemCom framework is summarized in Table \ref{tab:model}. For all convolutional layers in the network, the configuration is denoted as \emph{(filters, kernel size, strides)}, where \emph{filters} represents the number of output channels, \emph{kernel size} specifies the spatial or temporal receptive field, and \emph{strides} controls the downsampling factor.

\begin{table}[htbp]
	\centering
	\scriptsize 
	\setlength{\tabcolsep}{2pt} 
	\caption{Network Architecture of the Proposed Semantic Communication Framework}
	\label{tab:model}
	\begin{tabular}{cllc}
		\toprule
		& \textbf{Module} & \textbf{Layer} & \textbf{Configuration} \\
		\midrule
		\multirow{19}{*}{Encoder} 
		& & Conv2D+BN+ReLu & (64,3,1) \\
		& ResBlock & & (64,3,2) \\
		& ResBlock & & (128,3,2) \\
		& ResBlock & & (128,3,1) \\
		\cmidrule{2-4}
		& \makecell[l]{Noise\\Adaptation} & Dense+Sigmoid & 128 \\
		\cmidrule{2-4}
		& \multirow{5}{*}{\makecell[l]{Scenario\\Adaptation}} 
		& Conv1D+ReLu & (16,5,2) \\
		& & Conv1D+ReLu & (32,5,2) \\
		& & AveragePooling & --- \\
		& & Dense+ReLu & 64 \\
		& & Dense+Sigmoid & 64 \\
		\cmidrule{2-4}
		& & Conv2D & ($\frac{N_{sf} \times N_{st} \times N_k}{64}$,3,1) \\
		& & Flatten & --- \\
		& & Dense+Tanh & $N_{sf} \times N_{st} \times N_k \times 2$ \\
		\cmidrule{2-4}
		& \makecell[l]{OFDM\\Map} & & \\
		\cmidrule{2-4}
		& \multirow{3}{*}{\makecell[l]{NN\\Precoding}} 
		& Conv2D+BN+ReLu & (64,3,1) \\
		& & Conv2D+BN+ReLu & (64,3,1) \\
		& & Conv2D & (4,1,1) \\
		\cmidrule{2-4}
		& \makecell[l]{Power\\Normalization} & & \\
		\midrule
		\multirow{16}{*}{Decoder}
		& \multirow{3}{*}{\makecell[l]{Pilot-Guided\\Attention}} 
		& SepConv2D+BN+ReLu & (64,3,1) \\
		& & SepConv2D+BN+ReLu & (64,3,1) \\
		& & SepConv2D+Sigmoid & ($\frac{N_f \times N_t \times N_k}{64}$,3,1) \\
		\cmidrule{2-4}
		& \multirow{5}{*}{\makecell[l]{Classification\\Net}} 
		& SepConv2D+ReLu+MaxP & (16,3,1) \\
		& & SepConv2D+ReLu+MaxP & (32,3,1) \\
		& & SepConv2D+ReLu+MaxP & (64,3,1) \\
		& & Flatten & --- \\
		& & Dense & 128 \\
		& & Dense & number of classes \\
		\cmidrule{2-4}
		& \multirow{7}{*}{\makecell[l]{Restoration\\Net}} 
		& Conv2D+ReLu & (128,1,1) \\
		& & Upsampling & (2,2) \\
		& & SepConv2D+BN+ReLu & (64,3,1) \\
		& & Upsampling & (2,2) \\
		& & SepConv2D+BN+ReLu & (32,3,1) \\
		& & SepConv2D & (16,3,1) \\
		& & Conv2D+Sigmoid & (3,3,1) \\
		\bottomrule
	\end{tabular}
\end{table}

\vspace{-1mm}
\subsection{Asymmetric Encoder}
The asymmetric encoder is deployed at the BS, where sufficient computational resources are available. Its objective is to extract high level semantic features from input images and map them into channel symbols suitable for MU-MIMO OFDM transmission. 

Specifically, the encoder adopts a deep residual convolutional architecture to enhance representation capability while maintaining stable training behavior. Given an input image, a convolutional stem first extracts low level visual features, followed by a stack of residual blocks that progressively downsample the spatial resolution and increase the feature dimensionality. This design enables the encoder to capture both global structure and fine-grained semantic information.

To achieve scenario-aware semantic adaptation, the encoder explicitly incorporates CSI and SNR information. The frequency-domain channel response is first transformed into a magnitude representation, which captures both large-scale path loss and frequency-selective fading characteristics. A dedicated CSI feature extraction network processes this representation to generate a scenario scaling factor. In parallel, the noise variance is processed through a fully connected layer. The resulting CSI and SNR scaling factor are jointly applied to scale intermediate semantic features, enabling the encoder to adaptively emphasize essential semantic under adverse channel conditions.

Finally, the scaled features are mapped to complex-valued symbols and reshaped onto the OFDM resource grid, where known pilot symbols $\mathbf{Y}_p$, shared by both the transmitter and the receiver, are explicitly inserted at predefined frequency locations. A power normalization step is applied to satisfy the average transmit power constraint. This asymmetric design allows the BS to perform sophisticated semantic and channel-aware processing, while significantly reducing the burden on the UE.

\vspace{-3mm}
\subsection{Precoding Network}
Standard ZF precoding faces severe numerical instability in realistic fading channels due to the inversion of ill-conditioned matrices $(\mathbf{HH}^H)^{-1}$, which often leads to gradient explosion during training. Although RZF mitigates this singularity by introducing a regularization term, traditional RZF schemes typically rely on fixed or statistically derived regularization factors that are suboptimal for semantic transmission tasks. Moreover, as a linear operator, standard RZF cannot compensate for the non-linear distortions inherent in deep semantic features.

\begin{figure}[h]
	\centering
	\includegraphics[width=2.5in]{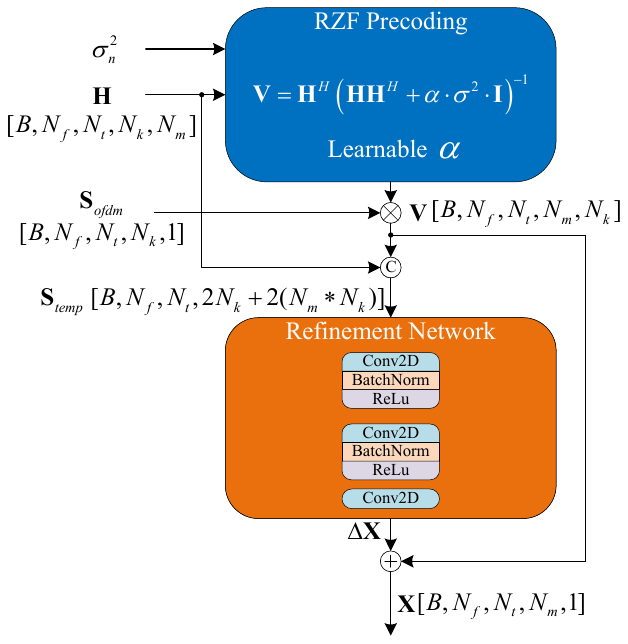}
	\caption{Architecture of the proposed hybrid learnable residual RZF precoding module.}
	\label{fig_4}
\end{figure}

To address these limitations, we propose a hybrid learnable residual RZF precoder as shown in Fig. \ref{fig_4}. By incorporating the RZF structure as a differentiable module with a learnable regularization parameter, our method inherits the numerical stability of RZF while enabling data-driven optimization. Furthermore, we introduce a residual refinement network to capture the non-linear spatial and semantic correlations that linear precoder overlook.

The input to the precoder comprises the semantic feature tensor extracted by the encoder, denoted as $\mathbf{S}_{\mathrm{ofdm}} \in \mathbb{C}^{B \times N_f \times N_t \times N_k}$, and the channel frequency responses $\mathbf{H}\in \mathbb{C}^{B \times N_f \times N_t \times N_k \times N_m }$, where $B$ is the batch size. In our implementation, matrix operations for RZF are applied to the last two tensor dimensions, with broadcasting enabling parallel processing. For neural refinement, the time and frequency dimensions are treated as spatial dimensions, and the remaining dimensions are merged into the channel axis. The real and imaginary parts are concatenated along the channel dimension.

In the first stage, we compute a coarse precoded signal using a differentiable implementation of RZF. Unlike standard RZF which uses a fixed regularization factor, we introduce a learnable parameter $\alpha$ to dynamically adjust the regularization strength based on the global interference level. The coarse precoding matrix $\mathbf{V}$ is computed as:
\begin{equation} 
	\mathbf{V} = \mathbf{H}^H \left(\mathbf{H}\mathbf{H}^H + \alpha \cdot \sigma^2 \cdot \mathbf{I} \right)^{-1}, 
\end{equation}
where $\mathbf{I}$ is the identity matrix. By making $\alpha$ trainable, the network gains the ability to adaptively optimize the RZF precoding process.

The linear RZF assumes independent Gaussian symbols, ignoring the spatial and semantic correlations inherent in image features. To capture these deficiencies, we employ a residual refinement network $\mathcal{F}_p(\cdot)$. The inputs to this network are the concatenation of the coarse precoded signal and the CSI. We flatten the spatial dimensions into the feature channel dimension to facilitate CNN processing. The network consists of two stacked convolutional layers with batch normalization and ReLU activation, followed by a convolutional layer. It predicts a residual correction term $\Delta X$:
\begin{equation} 
	\Delta \mathbf{X} = \mathcal{F}_p(\mathbf{S}_{temp}; \zeta).
\end{equation}
The final precoded signal is the superposition of coarse precoded signal and residual correction term:
\begin{equation} 
	\mathbf{X} = \mathbf{VS}_{ofdm} + \Delta \mathbf{X}. 
\end{equation}
Finally, $\mathbf{x}$ is reshaped back to the strictly required transmission format $[B, N_f, N_t, N_m, 1]$ and normalized to satisfy the total transmit power constraint $E[||\mathbf{X}||^2] \le P$.

\vspace{-1mm}
\subsection{Lightweight Decoder}
To enable deployment on computationally restricted UE, the decoder design must strike a balance between computational efficiency and semantic recovery performance. We propose a lightweight architecture that integrates a pilot-guided mechanism with Depthwise Separable Convolutions (DS-Conv) \cite{howard2017mobilenets}. Unlike standard convolutional layers which are computationally expensive, DS-Conv factorizes the operation into two distinct stages: a depthwise convolution that spatially filters each input channel independently, and a subsequent $1 \times 1$ pointwise convolution that linearly combines the resulting feature maps. This factorization significantly reduces the model size and floating point operations (FLOPs), allowing the UEs to perform deep semantic decoding with minimal energy consumption. 

\begin{figure}[h]
	\centering
	\includegraphics[width=2.5in]{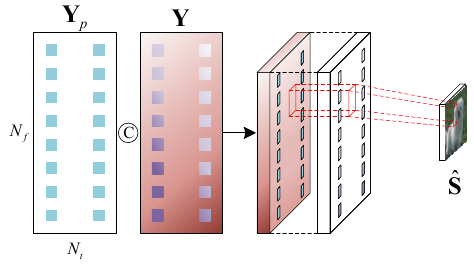}
	\caption{Illustration of the pilot insertion and pilot-guided fusion strategy}
	\label{fig_5}
\end{figure}

To robustly recover semantic information under fading channels without explicit channel estimation, we employ a pilot-guided fusion strategy as shown in Fig. \ref{fig_3}. We first construct a sparse reference pilot tensor, denoted as $\mathbf{Y}_p$, which contains the known transmitted symbols at allocated subcarriers and zeros at data positions, as shown in Fig. \ref{fig_5}. At the decoder input, the received signal $\mathbf{Y}$ is concatenated with $\mathbf{Y}_p$ along the channel dimension. This combined tensor is then processed by the initial convolutional layers. Since the convolutional kernels possess a receptive field covering both the distorted pilots in $\mathbf{Y}$ and the clean references in $\mathbf{Y}_p$, the network implicitly infers the CSI by analyzing the discrepancy between them. This learned knowledge is then applied to equalize neighboring semantic symbols within the same receptive field, enabling dynamic adaptation to varying channel conditions.

To further suppress noise and enhance feature reliability, the fused features are processed by an attention module to generate a weight map. These weights are applied via element multiplication with the input feature to reduce the impact of noise. To ensure that the deep decoder retains full access to the original reference information throughout the network, the weighted features are concatenated once more with the pilot reference before entering the main backbone. This design explicitly leverages the known pilot sequences to guide the feature recovery process, effectively performing implicit channel equalization and noise suppression.

The refined feature maps are subsequently processed by the lightweight backbone, which adopts up-sampling layers to progressively recover spatial resolution. Finally, the network splits into two lightweight task heads: a reconstruction head that maps latent features back to the pixel space to recover the original image, and a classification head that utilizes global average pooling to extract high-level semantic descriptors for image classification. This multi-task design ensures that the system preserves both perceptual details and semantic meaning.

\vspace{-1mm}
\section{Experimental Results}\label{result}
\begin{table}[htbp]
	\caption{Simulation Parameters}
	\label{tab:sim_params}
	\centering
	\begin{tabular}{lcc}
		\toprule
		\textbf{Parameter} & \textbf{Symbol} & \textbf{Value} \\
		\midrule
		\multicolumn{3}{l}{\textit{OFDM Settings}} \\
		Number of OFDM Symbols & $N_t$ & 14 \\
		Number of Subcarriers & $N_f$ &  \{32,64,128\} \\
		Carrier Frequency & - & 2.6 GHz \\
		Subcarrier Spacing & - & 15 kHz \\
		Pilot Symbol Indices & - & [2, 11] \\
		Cyclic Prefix Length & - & 20 \\
		Code rate & $r$ & - \\
		Modulation order & $m$ & - \\
		\midrule
		\multicolumn{3}{l}{\textit{Base Station (BS) Configuration}} \\
		Number of BS Antennas & $N_m$ & 4 \\
		Antenna Array Geometry & - & $1 \times 2$ (Row $\times$ Col) \\
		Polarization & - & Dual (Cross) \\
		Antenna Element Pattern & - & 3GPP 38.901 \\
		\midrule
		\multicolumn{3}{l}{\textit{User Equipment (UE) Configuration}} \\
		Number of Users & $N_k$ & 4 \\
		Antennas per UE & - & 1 \\
		Polarization & - & Single (Vertical) \\
		Antenna Element Pattern & - & Omni-directional \\
		\bottomrule
	\end{tabular}
\end{table}

Datesets: For numerical evaluation, we employ the widely adopted CIFAR-10 dataset \cite{krizhevsky2009learning}. This dataset comprises a total of 60,000 RGB images, each with a spatial resolution of $32\times32$ pixels, uniformly distributed across 10 distinct semantic categories. The dataset is officially partitioned into a training set of 50,000 samples and a testing set of 10,000 samples. Given its diverse content and balanced class distribution, CIFAR-10 serves as a robust standard for benchmarking performance in joint image transmission and classification tasks.

\begin{figure}[ht]
	\centering
	\includegraphics[width=2.5in]{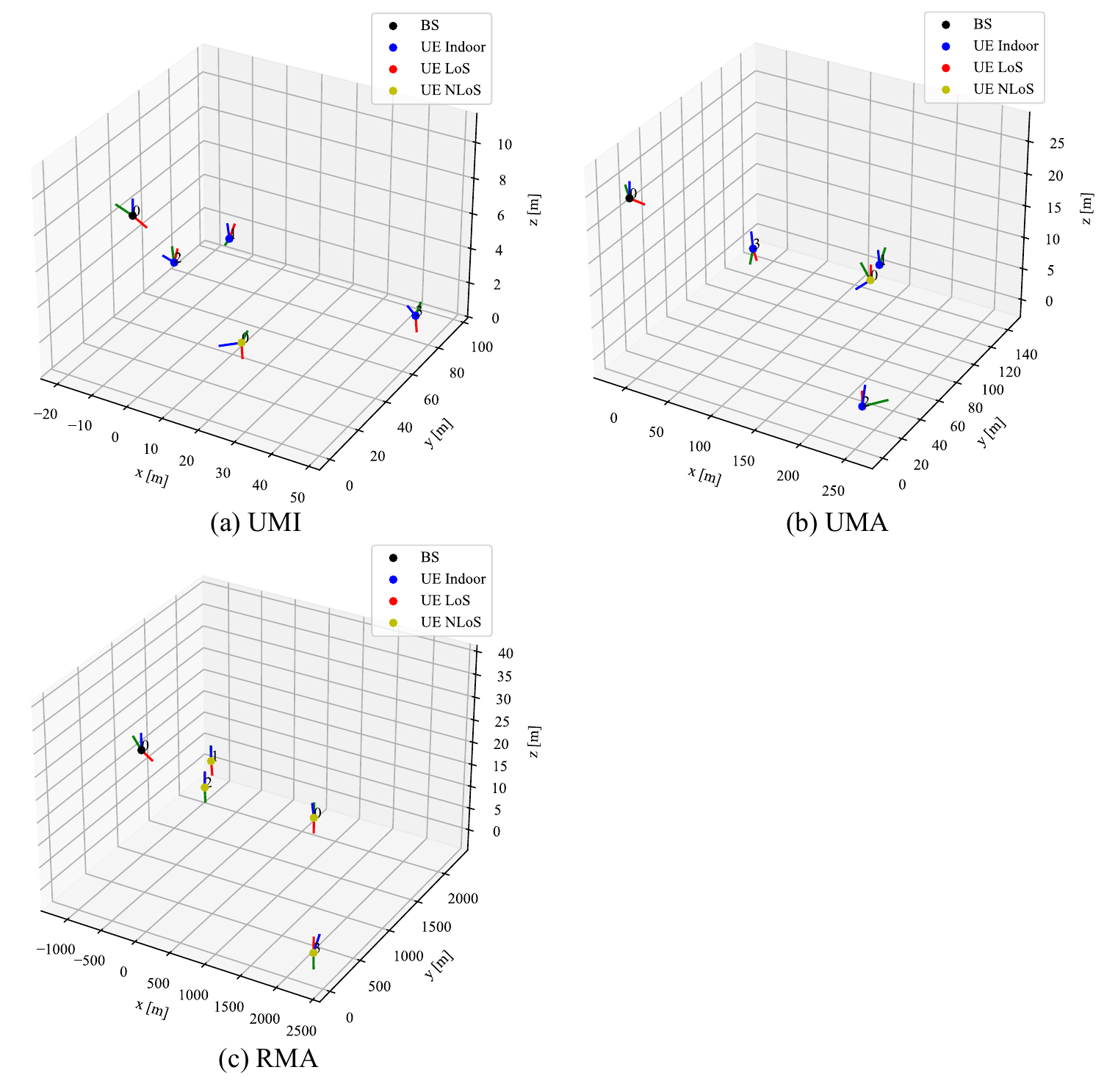}
	\caption{Spatial deployment of BS and UE in UMi, UMa, and RMa simulation environments}
	\label{3gpp}
\end{figure}

Communication System Simulation Setup: In this work, we implement a downlink MU-MIMO OFDM simulation environment based on the Sionna library\cite{sionna}. To comprehensively evaluate the generalization capability and robustness of the proposed framework, we employ three distinct 3GPP TR 38.901 channel models: UMi, UMa, RMa. These scenarios cover a wide range of path loss and delay spread conditions. An example of the spatial deployment is illustrated in Fig. \ref{3gpp},  while in the actual simulations the user locations and propagation geometry are randomly generated at each training. The detailed simulation parameters and system specifications are summarized in Table \ref{tab:sim_params}.

Training Detail: All experiments were conducted on a workstation equipped with an Intel Core i5-13600K CPU, an NVIDIA GeForce RTX 4070 Ti Super GPU, and 128 GB of RAM. The training was performed under Ubuntu 20.04 running on Windows Subsystem for Linux 2 (WSL2). For the semantic transmission network, the batch size was set to 64 and the learning rate was initialized at $1\times 10^{-3}$. The model was trained for 100 epochs using the Adam optimizer. The optimization objective is defined as a composite loss function, employing MSE for image reconstruction and CE loss for semantic classification.

The classical benchmarks adopt the standard SSCC architecture, utilizing Better Portable Graphics (BPG)\cite{bpg} or Joint Photographic Experts Group (JPEG) for source compression followed by 5G New Radio (NR) Low-Density Parity-Check (LDPC) codes for channel error correction. To ensure valid transmission under strictly limited bandwidth resources, we implemented an adaptive rate control mechanism. Specifically, the maximum allowable number of information bits per transmission block, denoted as $B_{max}$, is dynamically determined by the available OFDM resource elements, the selected modulation order, and the LDPC code rate. This capacity constraint is formulated as:
\begin{equation}
	B_{max} = \lfloor N_{st}\times N_{sf} \times m \times r \rfloor
\end{equation}
Where $\lfloor \cdot \rfloor$ denotes the floor operator. Prior to transmission, the source image is compressed, and the resulting bitstream length is compared against $B_{max}$. If the bitstream exceeds this capacity limit, the compression quality is iteratively reduced until the size constraint is met. In the extreme case where the compressed bitstream remains larger than $B_{max}$ even at the lowest quality setting, the transmission is considered a failure, and the PSNR is recorded as 0 dB. For successful compressions, zero-padding is applied to match the required input block length. At the receiver, Least Squares channel estimation is first performed to acquire the channel state information. Subsequently, Linear Minimum Mean Squared Error equalization is applied to recover the transmitted symbols based on the estimated channel and noise variance. Finally, the equalized symbols are processed by a demapper to compute Log-Likelihood Ratios, which are then fed into the LDPC decoder to recover the information bits.

In addition to classical benchmarks, we consider DJSCC\cite{bourtsoulatze2019deep} as the primary deep learning baseline. DJSCC is a pioneering end-to-end SemCom framework that employs a convolutional autoencoder to jointly optimize source and channel coding. For a fair comparison, we adapted the original DJSCC architecture to the MU-MIMO OFDM environment, maintaining the same channel settings and bandwidth constraints as our proposed framework. 

To further validate the effectiveness of our specific design for neural precoding and the pilot-guided decoding, we implement several variants of our model for an ablation study. These variants allow us to disentangle the impact of each module and highlight their respective contributions to the overall performance:

\begin{itemize} 
\item \textbf{Ours (RZF):} In this variant, the transmitter employs the classical RZF precoding scheme instead of the proposed neural precoder, and the adaptive module is disabled. At the receiver, the pilot-guided attention mechanism is also removed.

\item \textbf{Ours (w/o Pilot):} This variant adopts the proposed neural network precoder at the transmitter together with the adaptive module. However, the pilot-guided attention mechanism at the receiver is disabled.

\item \textbf{Ours (Full):} The complete proposed framework, which integrates all designed modules, including the neural network precoder and the adaptive module at the transmitter, as well as the pilot-guided lightweight decoder at the receiver.

\end{itemize}

\vspace{-1mm}
\subsection{Performance of Traditional Transmission}
\begin{figure*}[ht]
	\centering
	\includegraphics[width=7in]{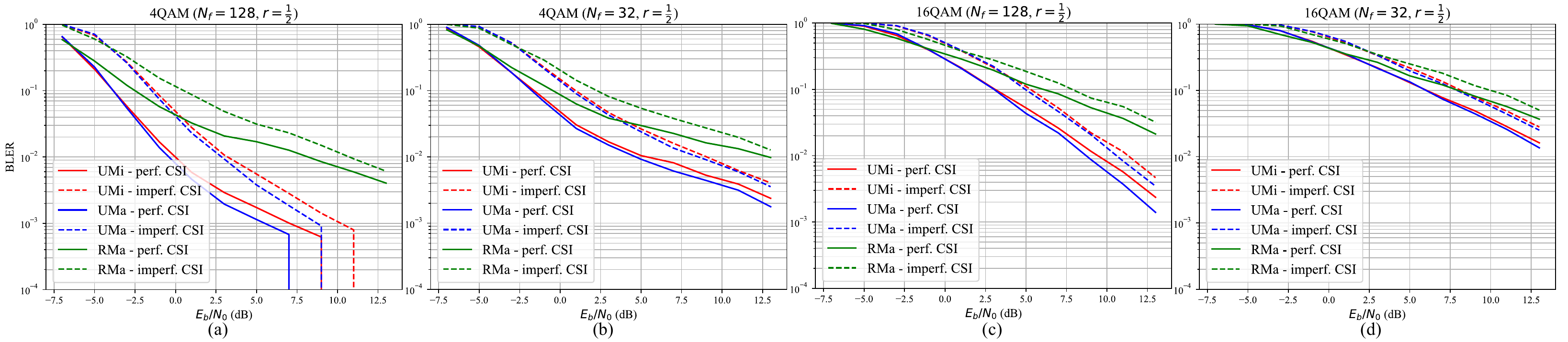}
	\caption{BLER Performance of 4×4 MU-MIMO Downlink Under Diverse 3GPP 38.901 Channel Models, Modulation Schemes and Subcarriers Numbers}
	\label{traditional_Bler}
\end{figure*}
Fig. \ref{traditional_Bler} illustrates the block error rate (BLER) performance of conventional digital transmission schemes under different modulation and coding configurations. The results are evaluated over multiple standardized propagation scenarios under both perfect and imperfect CSI assumptions. The vertical axis shows the BLER as a function of the SNR $E_b/N_0$.

Several important observations can be drawn from the figure. First, the system performance varies significantly across different propagation environments due to their distance fading characteristics. Across all modulation and code rate setting, the UMA scenario consistently exhibits the most robust performance, achieving the lowest BLER at equivalent SNR levels. This is attributed to the higher probability of LoS components in UMa environments. Conversely, RMA scenario demonstrates the poorest performance, due to larger delay spreads and more severe frequency-selective fading inherent to the RMa model in the simulated topology. The UMI scenario falls in between. 

Second, increasing the number of subcarriers from $N_f = 32$ to $N_f = 128$ consistently improves BLER performance across all channel models and modulation orders. This improvement is mainly attributed to enhanced frequency diversity, which mitigates deep fades and improves decoding reliability in frequency-selective channels. The gain is particularly evident in harsh propagation environments, where larger $N_f$ leads to a more pronounced BLER reduction at moderate-to-high SNR regimes.

Third, the modulation order has a significant impact on system robustness. Compared with 4QAM, 16QAM exhibits noticeably higher BLER under the same coding rate $r = 1/2$, especially in low-to-medium SNR regions. This behavior reflects the classical tradeoff between spectral efficiency and reliability in traditional communication systems, where higher-order modulation schemes are more sensitive to noise and channel impairments.

Finally, imperfect CSI leads to a degradation in BLER performance for all configurations. The degradation caused by channel estimation errors is substantial. For instance. in the 4QAM $N_f=128$ case, achieving a target BLER of $10^{-1}$ under imperfect CSI requires an additional SNR margin of approximately 2-3 dB compared to the perfect CSI case. 

\begin{figure*}[ht]
	\centering
	\includegraphics[width=7.0in]{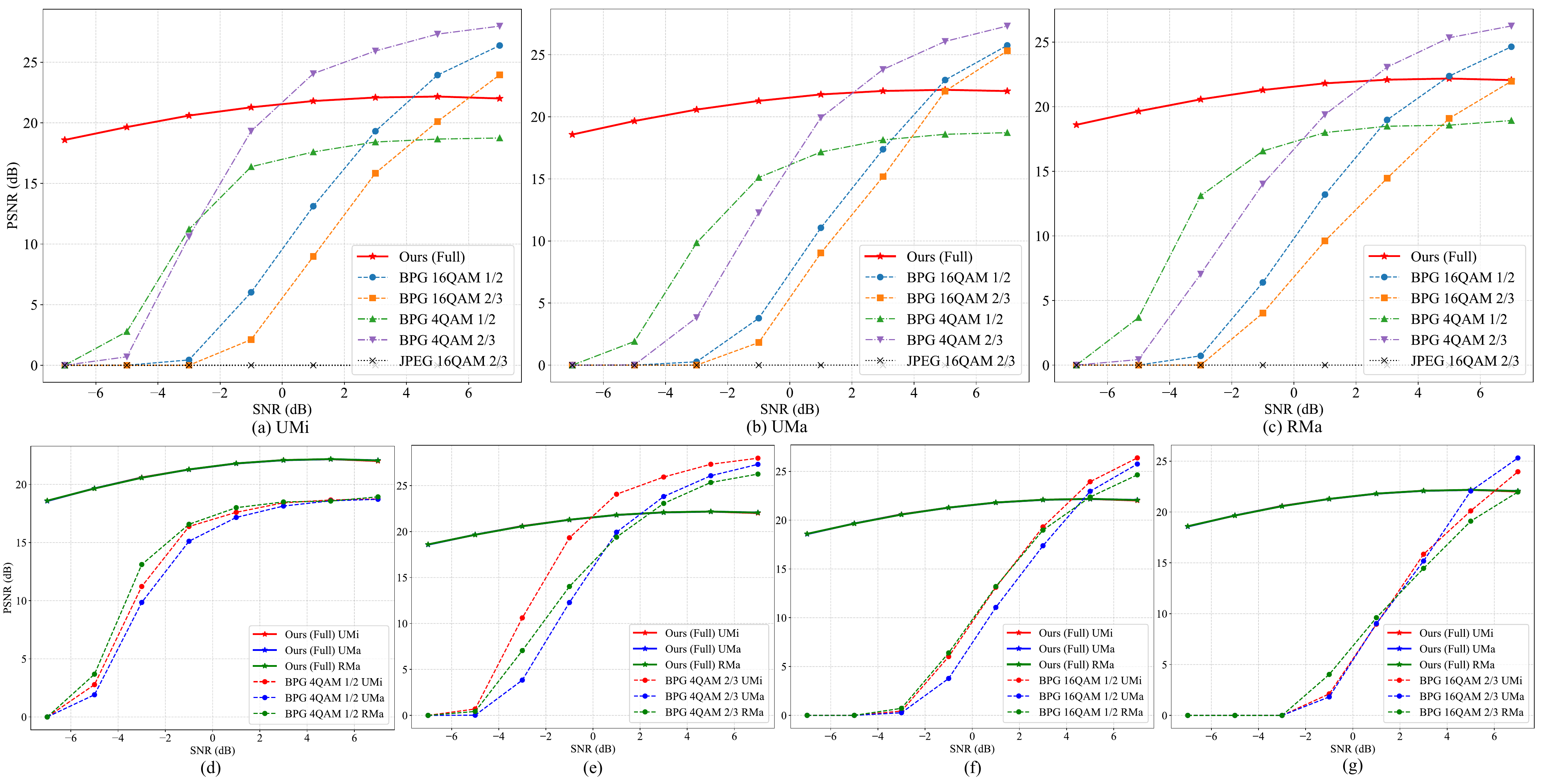}
	\caption{Comparison of SSCC image transmission performance under different scenario ($N_f=32$)}
	\label{Nf32}
\end{figure*}

Fig. \ref{Nf32} illustrates the PSNR performance of the proposed framework compared to the classical SSCC benchmarks under different modulation and coding rate. The results highlight several critical insights regarding robustness, scenario adaptability, and the trade-off between capacity and reliability.

\subsubsection{Superiority in Low-SNR Regime} As observed in the figures, the proposed method exhibits significant superiority in the low-SNR regime. For instance, in the UMi scenario, our method achieves a decent PSNR of approximately 18.6 dB at the lowest SNR point, whereas the SSCC schemes yield a PSNR of 0 dB due to decoding failures. This demonstrates that the proposed end-to-end design learns robust feature representations that gracefully degrade with channel quality.

\subsubsection{Scenario Robustness} The proposed method demonstrates remarkable consistency across different channel environments, with the performance curves for UMi, UMa, and RMa scenarios being tightly clustered. This stability is primarily attributed to the inherent generalization capability of the data-driven deep learning paradigm. Unlike model-based approaches that rely on explicit channel assumptions, the end-to-end training enables the transceiver to learn universal semantic representations that are resilient to diverse fading statistics. In contrast, the SSCC benchmarks are sensitive to specific channel conditions. Specifically, the SSCC performance degrades most severely in the RMa scenario compared to UMi and UMa. This aligns with the BLER performance observed in Fig. \ref{traditional_Bler}.

\subsubsection{Correlation with BLER} The PSNR trends of the SSCC methods in Fig. \ref{Nf32}(f) are intrinsically linked to the BLER results shown in Fig. \ref{traditional_Bler}(d) under the same simulation settings. In digital transmission systems, image reconstruction depends on the successful decoding of the entire data block, resulting in an essentially binary recovery behavior. Consequently, as the BLER decreases with improving SNR, the probability of successful image recovery increases, leading to the observed rise in average PSNR.

\subsubsection{Trade-off between Modulation and Coding} The Fig. \ref{Nf32} from (d) to (g) reveal the critical impact of different modulation and coding configurations on transmission performance. Specifically, the 4QAM 1/2 scheme shown in Fig. 8(d) exhibits the strongest noise resilience, enabling successful decoding at lower SNRs compared to other schemes. However, this robustness comes at the cost of low spectral efficiency, which severely restricts the maximum number of transmittable information bits $B_{max}$. Consequently, the source encoder is forced to apply excessive compression, resulting in a low PSNR of approximately 18.75 dB even under 7dB SNR channel conditions. As the modulation order increases to 16QAM and code rates rise to 2/3 (moving from Fig. \ref{Nf32}(d) to Fig. \ref{Nf32}(g)), this noise resilience progressively deteriorates, evidenced by the cliff point shifting toward higher SNR regions. Although higher-order modulation and higher coding rates allow more information bits to be transmitted, thereby reducing source compression distortion, but the corresponding loss in noise robustness prevents these schemes from fully exploiting favorable channel conditions. As a result, the PSNR does not increase proportionally with the available transmission rate, revealing a fundamental inefficiency in SSCC systems when operating under practical fading channels.

\subsubsection{Optimal Balance} Among the evaluated traditional baselines, the 4QAM scheme with a 1/2 code rate achieves the best trade-off between noise robustness and reconstruction quality. It can successfully decode at moderate SNR while reaching PSNR exceeding 26 dB under 7dB SNR. As a result, it outperforms both the overly compressed 4QAM 1/2 configuration and the more fragile 16QAM 2/3 scheme.

\subsection{Performance of Our Framework}

\begin{figure*}[ht]
	\centering
	\includegraphics[width=7.0in]{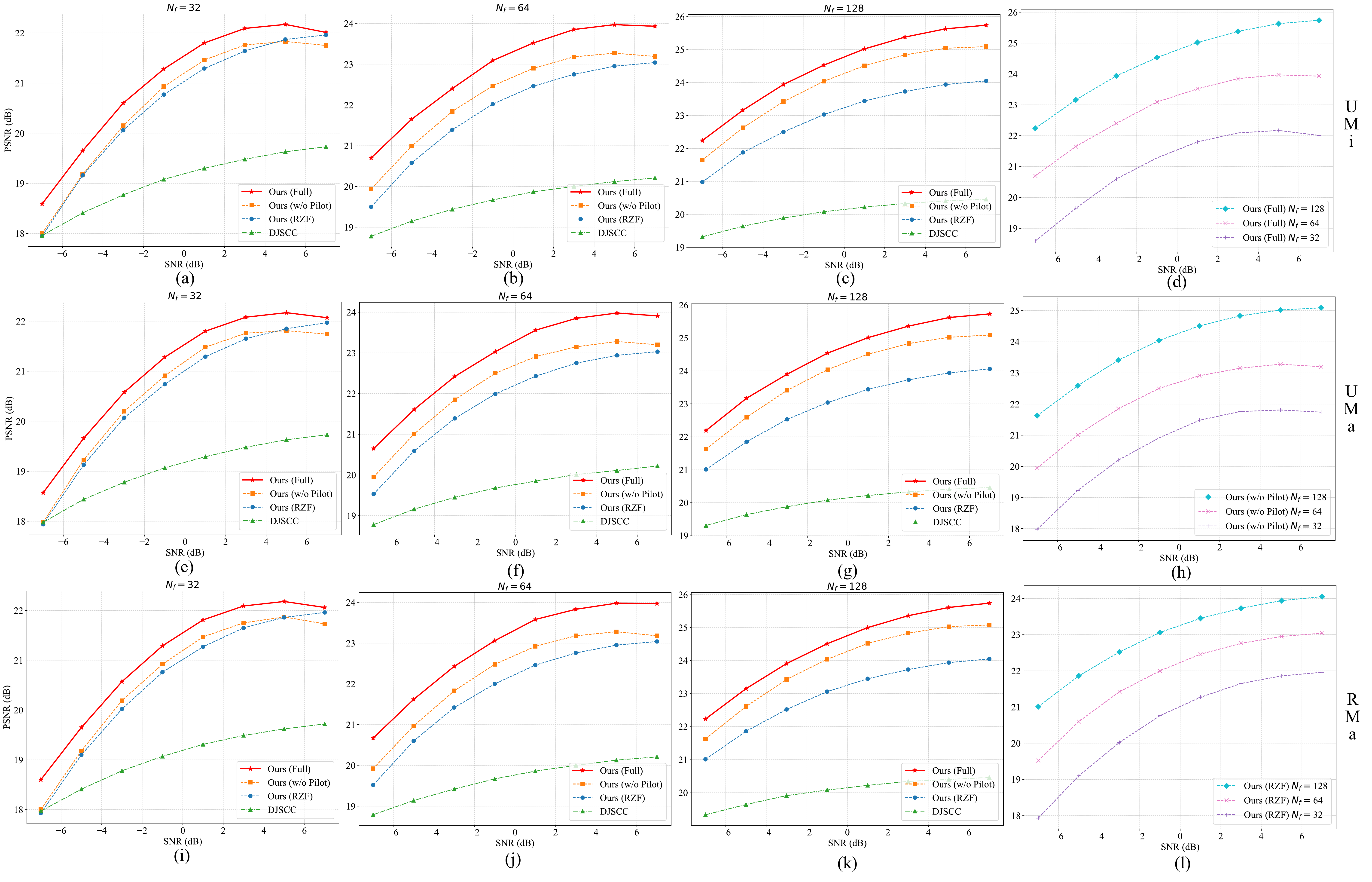}
	\caption{PSNR performance comparison of the proposed framework against the DJSCC baseline and ablation variants (Ours (w/o Pilot), Ours (RZF)) across UMi (top row), UMa (middle row), and RMa (bottom row) scenarios. Columns 1-3 illustrate performance under fixed bandwidth allocations ($N_f \in \{32, 64, 128\}$), while the last column (d,h,l) demonstrates the impact of bandwidth scaling on the reconstruction quality.}
	\label{ours_vs_djscc}
\end{figure*}

Fig. \ref{ours_vs_djscc} illustrates the PSNR performance of the proposed framework and its ablation variants compared to the DJSCC baseline \cite{bourtsoulatze2019deep}. The evaluation covers three channel scenarios  and three available bandwidth resources. Several key observations can be drawn from the results:

\subsubsection{Superiority over Baseline} First and foremost, the proposed framework exhibits a substantial performance advantage over the DJSCC baseline across all tested scenarios. As shown in the figure, DJSCC saturates early, with its peak PSNR hovering around 20.5 dB even with maximum bandwidth usage. Even our most basic variant employing RZF precoding (Ours (RZF)) significantly outperforms DJSCC by approximately 3.5 dB. This performance gap highlights a fundamental architectural mismatch when applying JSCC models to complex multi-user systems.

\begin{figure}[h]
	\centering
	\includegraphics[width=2.5in]{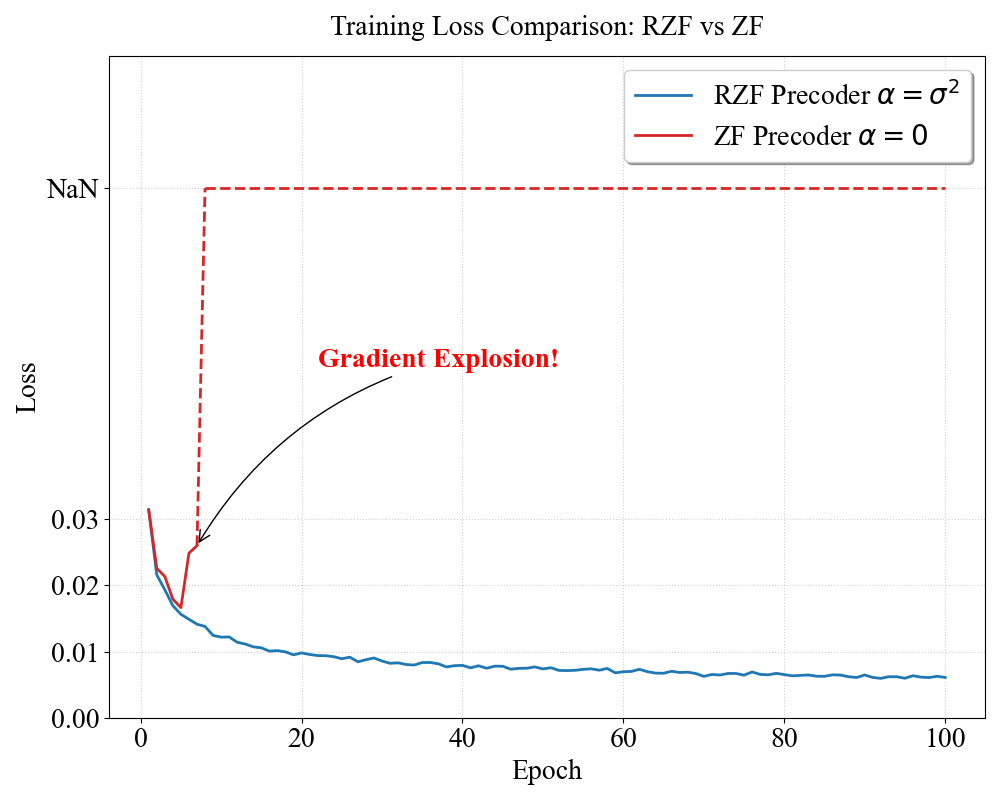}
	\caption{The training loss comparison between RZF and ZF precoder.}
	\label{rzf_vs_zf_loss}
\end{figure}

\subsubsection{Impact of Neural Precoding} First, we address the critical challenge of training stability in end-to-end MU-MIMO systems. Fig. \ref{rzf_vs_zf_loss}  illustrates the training loss trajectories of different precoding strategies. When the regularization factor is set to $\alpha=0$, the RZF scheme degenerates into the standard ZF precoder. As evidenced by the red dashed curve, the training collapses rapidly within the first few epochs, resulting in Not a Number (NaN) loss values. This numerical instability stems from the matrix inversion operation $(\mathbf{HH}^H)^{-1}$. During the initial training phases or under deep fading conditions, the channel matrix often becomes ill-conditioned or singular. This singularity generates astronomically large values during the forward pass, triggering severe gradient explosion during backpropagation and causing the learning process to fail. In contrast, our proposed neural precoder effectively bypasses these numerical pitfalls, ensuring robust and smooth convergence.

Regarding the signal reconstruction performance, the advantage of the proposed neural precoding module is evident when comparing Ours (w/o Pilot) against Ours (RZF). In high-bandwidth regimes ($N_f=128$), the neural precoder demonstrates clear superiority. For example, in the Umi scenario, Ours (w/o Pilot) achieves a maximum PSNR of 25.09 dB, surpassing Ours (RZF) by approximately 1 dB. This gain suggests that the neural precoder successfully learns a non-linear mapping strategy that is more effective than linear RZF in preserving high-dimensional semantic features against channel impairments. In low-bandwidth regimes ($N_f=32$), the performance gap narrows, with Ours (RZF) performing comparably to Ours (w/o Pilot). This implies that when semantic information is extremely compressed, the bottleneck shifts towards source coding loss, making the neural precoding less impactful compared to the RZF baseline.

\subsubsection{Effectiveness of Pilot-Guided Attention} The contribution of the Pilot-Guided Attention of decoder is evidenced by comparing Ours (w/o Pilot) with Ours (Full). The full model consistently achieves the highest performance across all SNR levels and channel configurations. By explicitly leveraging the reference pilot symbols via the proposed attention mechanism, Ours (Full) gains an additional improvement of approximately 0.7dB over the blind decoding version at high-bandwidth regimes ($N_f=128$). Even in the low-bandwidth regimes ($N_f=32$) case, the pilot guidance provides a stable gain, rising from 21.75 dB to 22.01 dB. This confirms that providing the decoder with an explicit reference effectively helps the network to implicitly equalize the channel and suppress residual noise, leading to sharper image reconstruction.

\subsubsection{Bandwidth Scalability} Comparing the columns from left ($N_t=32$) to right ($N_t=128$), we observe a positive correlation between available bandwidth and reconstruction quality. With $N_t=32$ (Fig. \ref{ours_vs_djscc} (a)), the peak PSNR is limited to $\approx$ 22 dB due to the severe compression ratio required to fit features into fewer subcarriers. By increasing the resource block to $N_t=128$ (Fig. \ref{ours_vs_djscc} (c)), the system can transmit richer semantic information, elevating the peak PSNR to nearly 26 dB. Crucially, our framework maintains its superiority over baselines regardless of the bandwidth constraint.

\begin{figure*}[ht]
	\centering
	\includegraphics[width=7.0in]{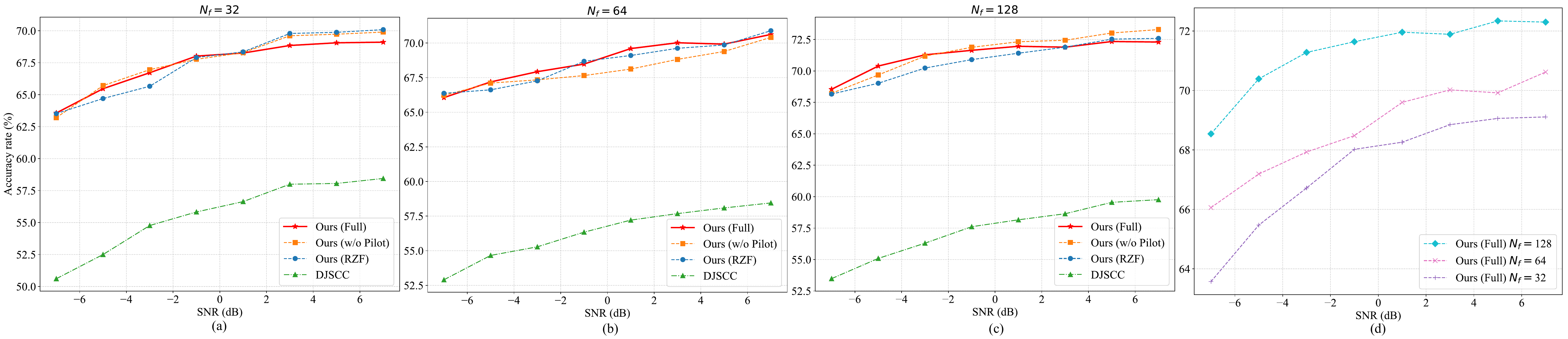}
	\caption{Classification accuracy (\%) versus SNR on the CIFAR-10 dataset in the UMi scenario. (a)-(c) Comparison of the proposed framework against DJSCC and ablation variants under fixed bandwidth allocations of $N_f=32$, $N_f=64$ and $N_f=128$, respectively. (d) Impact of bandwidth scaling on the classification accuracy of the proposed full model.}
	\label{ours_vs_djscc_cla}
\end{figure*}

In addition to image reconstruction, we evaluate the system's capability in semantic understanding tasks, specifically image classification on the CIFAR-10 dataset. Fig. \ref{ours_vs_djscc_cla} illustrates classification accuracy under the UMi scenario across varying bandwidth conditions. Consistent with the reconstruction results, the proposed framework demonstrates a substantial advantage over the DJSCC baseline in classification tasks. As shown in Fig. \ref{ours_vs_djscc_cla}(c), our method achieves a classification accuracy of approximately 72.3\% at 7dB SNR, whereas DJSCC saturates around 59.8\%. 

Comparing the variants of our proposed framework, we observe that all configurations achieve consistently high classification accuracy across the tested SNR range. For instance, in Fig. \ref{ours_vs_djscc_cla}(c), the three variants converge to a similar accuracy plateau of approximately 72\%-73\%, this phenomenon can be attributed to the inherent nature of semantic classification tasks. Unlike pixel-level reconstruction which demands precise waveform recovery, classification relies on high-level abstract features. These coarse-grained semantic features are naturally more resilient to channel distortions. Consequently, our robust encoder and neural precoder alone are sufficient to preserve the discriminative information required for accurate classification, ensuring reliable machine intelligence even without the additional refinement from the pilot-guided mechanism.

Fig. \ref{ours_vs_djscc_cla}(d) clearly depicts the correlation between available bandwidth and semantic accuracy. With limited resources ($N_f=32$), the accuracy is capped at approximately 69\%. Increasing the bandwidth to $N_f=128$ unlocks the potential for higher accuracy, reaching up to 72.3\%. However, it is worth noting that the marginal gain in classification accuracy from $N_f=64$ to $N_f=128$  is smaller compared to the gain in reconstruction PSNR (Fig. \ref{ours_vs_djscc}). This indicates that the semantic features needed for classification are highly compressible, and our lightweight encoder successfully captures the majority of semantic information even with moderate bandwidth allocation.

The results validate the effectiveness of our multi-task learning objective. By jointly optimizing for reconstruction and classification, the encoder learns a versatile latent representation that is both visually faithful and semantically meaningful. The system successfully achieves dual objectives within a unified transmission framework. It transmits images while ensuring the accurate conveyance of semantic information.

\begin{figure*}[ht]
	\centering
	\includegraphics[width=7.0in]{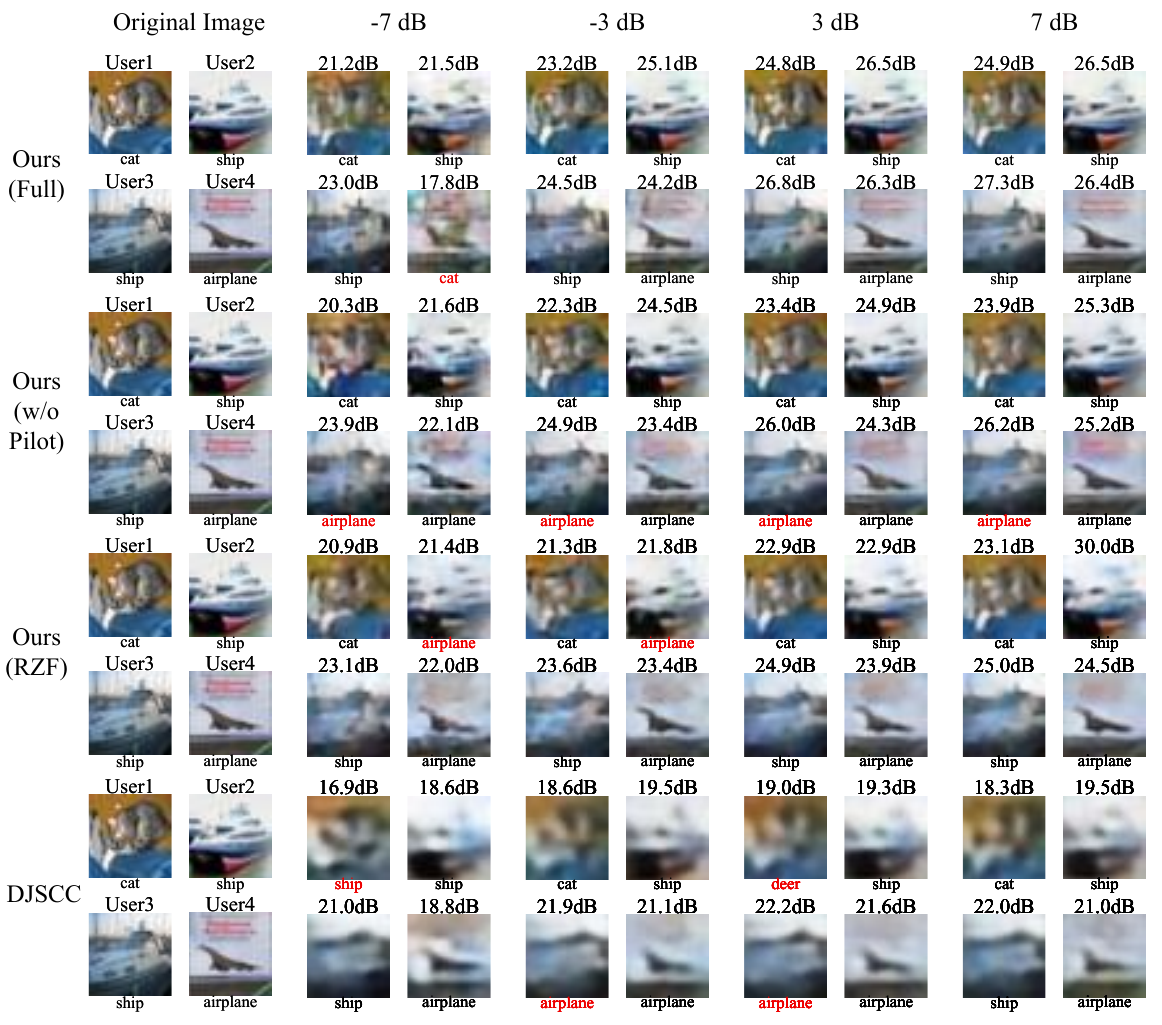}
	\caption{Visual quality comparison of reconstructed images and classification predictions in the downlink MU-MIMO system (UMi scenario $N_f=128$). The figure displays the recovered images, corresponding PSNR values, and predicted semantic labels across varying SNR levels. Red text indicates incorrect classification predictions.}
	\label{Visualization}
\end{figure*}

To provide an intuitive assessment of the semantic recovery performance, Fig. \ref{Visualization} visualizes the reconstructed images and classification results for four spatially multiplexed users under varying channel conditions. The proposed Ours (Full) framework demonstrates superior capability in preserving structural details and visual sharpness compared to the baselines. Even at the extremely low SNR of -7 dB, the images reconstructed by our method remain visually recognizable. In contrast, the images recovered by DJSCC (Bottom Row) suffer from severe blocking artifacts and blurring. This visual degradation confirms that the standard DJSCC architecture fails to effectively adapt to MU-MIMO systems, leading to a collapse of semantic features in the latent space.

Furthermore, the benefits of our architectural innovations are evident when comparing the ablation variants. While both Ours (w/o Pilot) and Ours (RZF) manage to convey the basic semantic content, they exhibit noticeable limitations in visual fidelity. As observed in the 3 and 7 dB, the images reconstructed by Ours (w/o Pilot) appear slightly oversmoothed and lack high-frequency texture details compared to the Full model. For instance, the intricate features of the ``cat'' (User 1) and the structural edges of the ``ship'' (User 2) are much sharper in the Full model, reflected in a consistent PSNR gain of approximately 1.0-1.5 dB. This confirms that the explicit injection of pilot symbols enables the decoder to perform fine-grained channel equalization, recovering subtle image details that are otherwise lost in blind decoding. Similarly, the Ours (RZF) variant, which relies on linear precoding, struggles to fully mitigate the non-linear distortion and residual multi-user interference, particularly at lower SNRs. In contrast, the full framework based on the neural precoder can effectively orthogonalizes user streams in the semantic domain, ensuring cleaner feature representations.

\begin{table*}[htbp]
	\centering
	\caption{Computational Complexity Comparison}
	\label{tab:complexity}
	\resizebox{\textwidth}{!}{%
		\begin{tabular}{lccccccccc}
			\toprule
			& \multicolumn{3}{c}{Params ($\times 10^6$)} & \multicolumn{3}{c}{Memory Usage (MB)} & \multicolumn{3}{c}{Inference time (ms)} \\
			\cmidrule(lr){2-4} \cmidrule(lr){5-7} \cmidrule(lr){8-10}
			& encoder & precoder & decoder & encoder & precoder & decoder & encoder & precoder & decoder \\
			\midrule
			Ours(Full)      & 10.13 & 0.061 & 0.11  & 38.65 & 0.23 & 0.42 & 61.9  & 29.73 & 46.8  \\
			Ours(w/o Pilot) & 10.13 & 0.061 & 0.078 & 38.65 & 0.23 & 0.3  & 60.16 & 31.64 & 31.93 \\
			Ours(RZF)       & 10.10  & --    & 0.078 & 38.54 & --   & 0.3  & 51.05 & 60.06 & 32.71 \\
			DJSCC           & 6.35  & 0.061 & 7.44  & 24.26 & 0.23 & 28.4 & 18.33 & 32.17 & 34.28 \\
			\bottomrule
		\end{tabular}%
	}
\end{table*}
Table \ref{tab:complexity} summarizes the computational complexity of different framework components in terms of model parameters (in millions), memory usage (MB), and average inference time (ms) measured on 4070Ti Super. A key design goal of our framework is to shift the computational burden from hardware-limited UEs to the resourceful BS. Our lightweight decoder (Ours (Full)) requires only 0.11 M parameters and consumes 0.42 MB of memory. This represents a drastic reduction of approximately 98.5\% in parameter count compared to the symmetric DJSCC decoder (7.44 M parameters, 28.4 MB memory). This lightweight nature ensures that our semantic receiver can be easily deployed on edge devices with limited storage.  Conversely, our BS encoder is designed to be heavy (10.13 M parameters) to maximize feature extraction capability. While this increases the computational load at the transmitter, it is a justifiable trade-off given the abundant resources available at the BS.

Comparing Ours (Full) with Ours (w/o Pilot), we observe a slight increase in the encoder inference time. This additional 10ms latency is the cost of the scenario and SNR adaptation modules included in the full model. However, considering the significant performance gains in robustness and generalization, this overhead is acceptable for most semantic transmission scenarios.

A significant finding is the speed of the precoding stage. Under GPU acceleration, our neural precoder is notably faster than the traditional RZF algorithm. This is because the neural precoder relies on highly parallelizable matrix multiplications and convolutions, which are optimized for GPU hardware. In contrast, the classical RZF algorithm involves explicit matrix inversion, an operation that scales cubically with dimension and is computationally expensive to parallelize efficiently. This result demonstrates that neural precoding is not only more robust to numerical instability but also more computationally efficient in modern hardware implementations.

\section{Conclusion}\label{conclusion}
In this paper, we presented a comprehensive end-to-end learning framework for semantic image transmission in downlink MU-MIMO OFDM systems. Recognizing the limitations of traditional SSCC and existing symmetric deep learning models, we designed an asymmetric architecture that jointly optimizes semantic extraction, neural precoding, and signal reconstruction.

Our experimental results lead to several key conclusions. First, we designed an asymmetric autoencoder architecture that shifts the computational burden to the BS, enabling the UE to utilize lightweight decoders. This approach is suitable for hardware-limited IoT and mobile devices. Second, the proposed environment-aware encoding and neural precoding mechanisms effectively mitigate multi-user interference and frequency-selective fading, overcoming the early performance saturation observed in standard DJSCC benchmarks. Third, the integration of a pilot-guided attention mechanism at the receiver significantly enhances decoding fidelity. By leveraging explicit pilot references, the lightweight decoder can robustly recover semantic features even in fading subcarriers, offering a substantial PSNR gain over blind decoding approaches. Finally, compared to the classical digital baseline (BPG + 5G LDPC), our framework demonstrates remarkable robustness across diverse channel scenarios (UMi, UMa, RMa), achieving graceful performance degradation in the low-SNR regime while maintaining high semantic classification accuracy.

These findings confirm the potential of incorporating domain knowledge, such as OFDM resource structures and pilot signals, into deep learning pipelines. Future work will extend this framework to massive MIMO scenarios with massive antenna arrays and investigate dynamic semantic resource allocation for users with heterogeneous task requirements.

\bibliographystyle{IEEEtran}
\bibliography{IEEEabrv,mybib}

\newpage

\vfill

\end{document}